\documentclass[12pt,letterpaper]{article}
\usepackage[utf8]{inputenc}
\usepackage[spanish, es-tabla]{babel}
\usepackage[version=3]{mhchem}
\usepackage[journal=jacs]{chemstyle}
\usepackage{amsmath}
\usepackage{amsfonts}
\usepackage{amssymb}
\usepackage{makeidx}
\usepackage{xcolor}
\usepackage[stable]{footmisc}
\usepackage[section]{placeins}
\usepackage{longtable}
\usepackage{array}
\usepackage{xtab}
\usepackage{multirow}
\usepackage{colortab}
\usepackage{cancel}
\usepackage{graphicx}
\usepackage{lmodern}
\usepackage{fancyhdr}
\usepackage[left=4cm,right=2cm,top=3cm,bottom=3cm]{geometry}

\usepackage{titlesec}
\usepackage{enumitem}
\titleformat*{\section}{\bfseries\large}
\titleformat*{\subsection}{\bfseries\normalsize}

\usepackage{float}
\floatstyle{plaintop}
\newfloat{anexo}{thp}{anx}
\floatname{anexo}{Anexo}
\restylefloat{anexo}

\usepackage[margin=10pt,labelfont=bf]{caption}

\usepackage{todonotes}

\usepackage[colorlinks=true, 
            linkcolor = blue,
            urlcolor  = blue,
            citecolor = black,
            anchorcolor = blue]{hyperref}

\usepackage{listings}
\usepackage[ruled,linesnumbered]{algorithm2e}

\SetAlFnt{\small}
\SetAlCapFnt{\small}
\SetAlCapNameFnt{\small}
\SetAlCapHSkip{0pt}
\IncMargin{-\parindent}

\usepackage{fancyhdr} 
\usepackage{lastpage} 
\usepackage{extramarks} 

 \newcommand{\hmwkTitle}{Apuntes de Redes Neuronales Artificiales} 
 \newcommand{\hmwkAuthorName}{Dr. Juan Carlos Cuevas Tello} 
 \title{Apuntes de Redes Neuronales (arXiv}
 \title{
 	\begin{figure}[ht]
 		\includegraphics [scale=0.25] {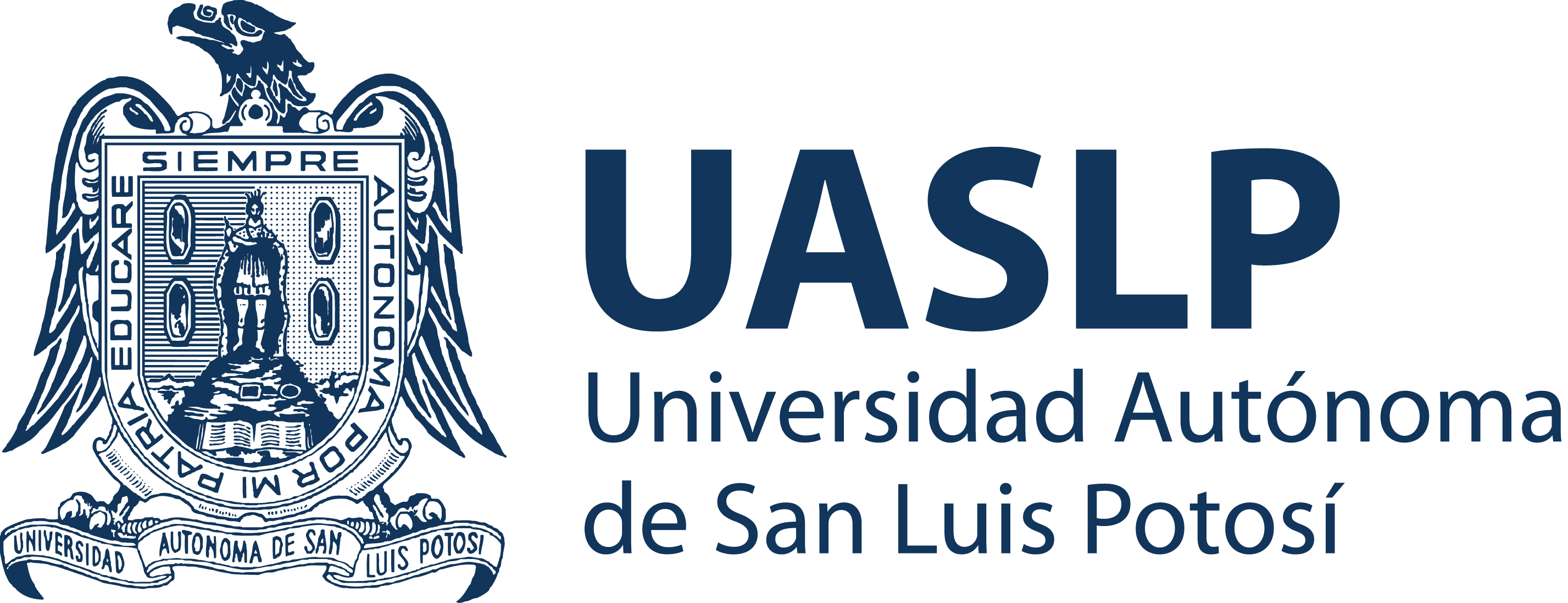}
        \hspace{1in}
         \includegraphics [scale=0.22] {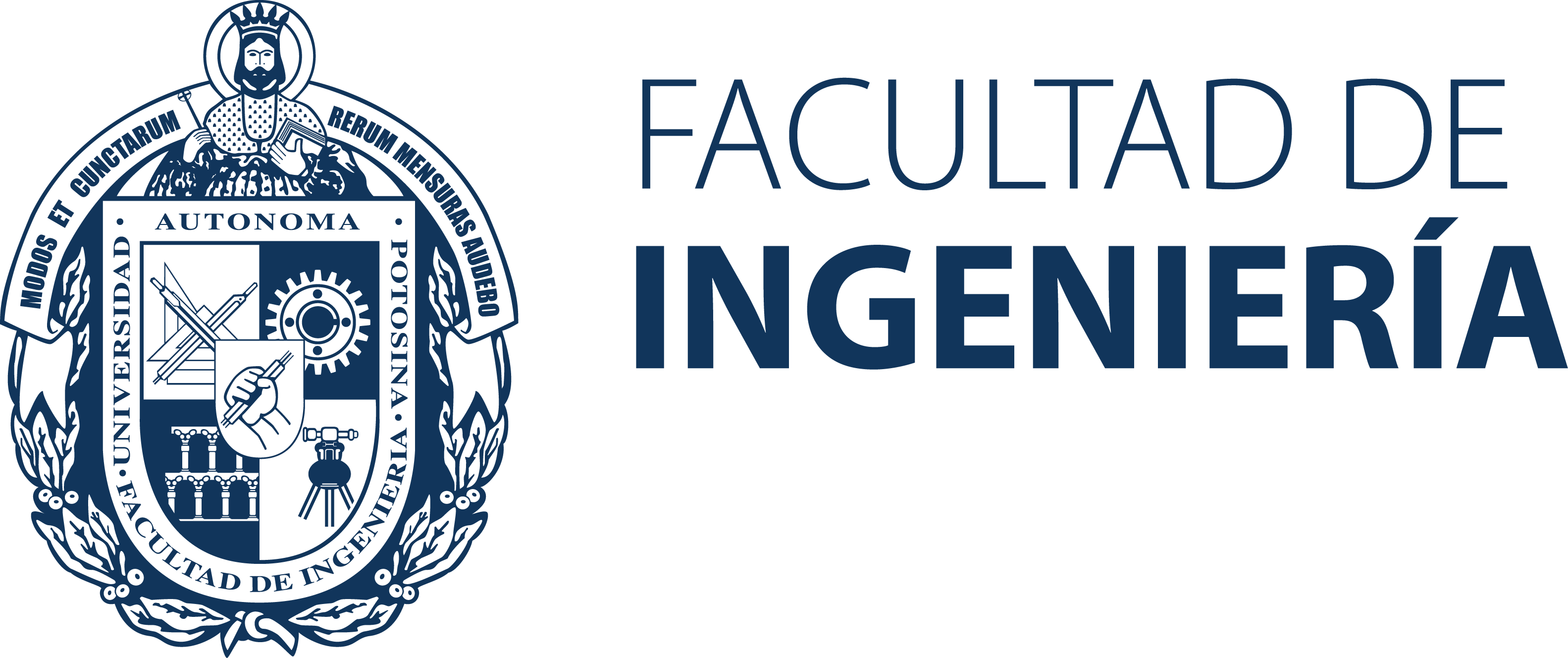}
 		\centering
 	\end{figure}	
 	\vspace{1.5in}	
 	\vspace{0.5in}		
 	\textmd{\textbf{ \hmwkTitle}}
    \textmd{\textbf{ Handouts for Artificial Neural Networks}}
  	\vspace{2.5in}
 } 
 \author{\textbf{\hmwkAuthorName}}
 \date{Agosto 2017 \\ \tiny{Última revisión: 4/Dic/2017}} 

\begin{document}
\maketitle
\setcounter{tocdepth}{1} 
\newpage
\tableofcontents
\newpage

\begin{abstract}
These handouts are designed for people who is just starting involved with the topic artificial neural networks. We show how it works a single artificial neuron (McCulloch \& Pitt model), mathematically and graphically. We do explain the delta rule, a learning algorithm to find the neuron weights. We also present some examples in MATLAB\textsuperscript{\textregistered{}}/Octave. There are examples for classification task for lineal and non-lineal problems. At the end, we present an artificial neural network, a feed-forward neural network along its learning algorithm backpropagation.

-----

Estos apuntes están diseñados para personas que por primera vez se introducen en el tema de las redes neuronales artificiales. Se muestra el funcionamiento básico de una neurona, matemáticamente y gráficamente. Se explica la Regla Delta, algoritmo de aprendizaje para encontrar los pesos de una neurona. También se muestran ejemplos en MATLAB\textsuperscript{\textregistered{}}/Octave. Hay ejemplos para problemas de clasificación, para problemas lineales y no-lineales. En la parte final se muestra la arquitectura de red neuronal artificial conocida como backpropagation.
\end{abstract}

\section{Introducción}

\begin{figure}[!ht]
\includegraphics[width=4in]{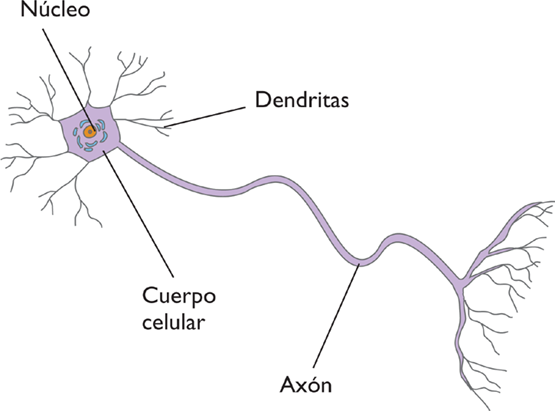}
\caption{Una célula, una neurona. Fuente: http://nauronas.blogspot.mx/}
\label{fig:1-1}
\end{figure}

Antes de hablar de redes neuronales artificiales es importante conocer como se constituye una célula, la neurona (Ver \ref{fig:1-1}).

Las Dendritas es la entrada de la información a la neurona, el Núcleo la procesa y la transmite a través del Axón. Las ramificaciones terminales del Axón se conectan a otras neuronas a través de un proceso conocido como \textit{sinápsis}. En la sinápsis entran en juego los neurotransmisores como la dopamina, serotonina entre muchos otros.

Se dice que hay alrededor de $10^{11}$ neuronas en el cerebro humano y cada neurona puede recibir información de $5,000$ a $15,000$ entradas de axones de otras neuronas \cite{Aleksander:1992:book}. 

\section{El modelo de McCulloch \& Pitts}
\begin{figure}
\includegraphics[width=3in]{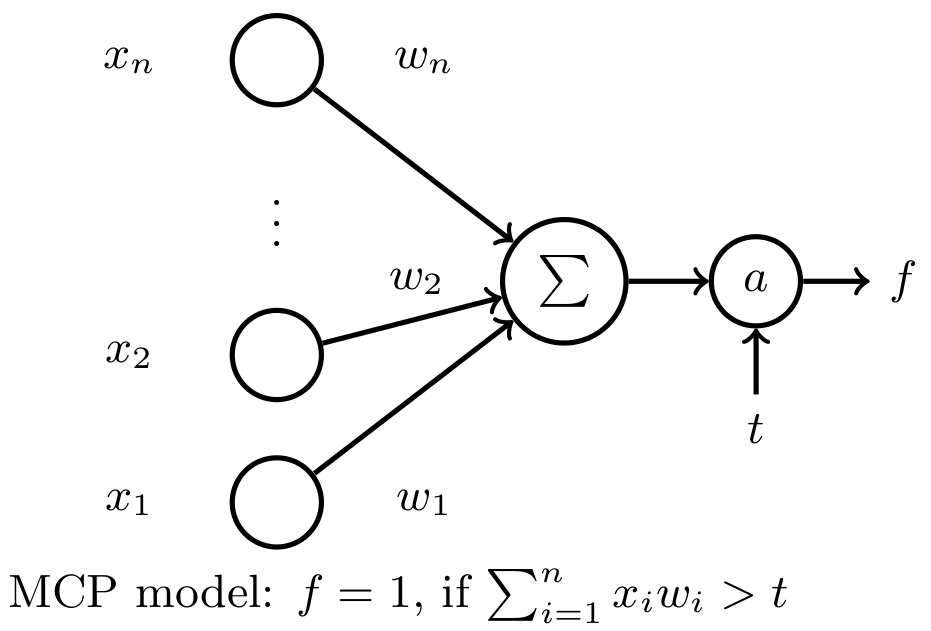}

\caption{Modelo McCulloch \& Pitts (MCP).}
\label{fig:mcp}
\end{figure}

La primera neurona artificial fue propuesta por Warren McCulloch y Walter Pitts en 1943 \cite{Aleksander:1992:book}, conocida como el modelo de McCulloch \& Pitts (MCP). El modelo se muestra en la \ref{fig:mcp}, se conoce también como TLU (Threshold Logic Unit) o LTU (Linear Threshold Unit) \cite{Aleksander:1992:book,Rojas:1996:book}.

En la \ref{fig:mcp}, las entradas $x_1, x_2,\dots, x_n$ simulan las dendritas y la salida $f$ la señal que viaja a través del axón.

Dependiendo de la función de activación $a$ la neurona puede ser lineal o no-lineal. En la \ref{fig:mcp} se muestra la versión lineal. Es decir, se genera un $f=1$ en la salida si $\sum_{i=1}^{n}x_i w_i > t$, en caso contrario la neurona genera $f=0$. Donde $w_1,w_2,\dots,w_n$ se conocen como pesos y $t$ como el umbral (threshold).

\subsection{Ejemplo: 2 entradas}
\begin{figure}
\includegraphics[width=4in]{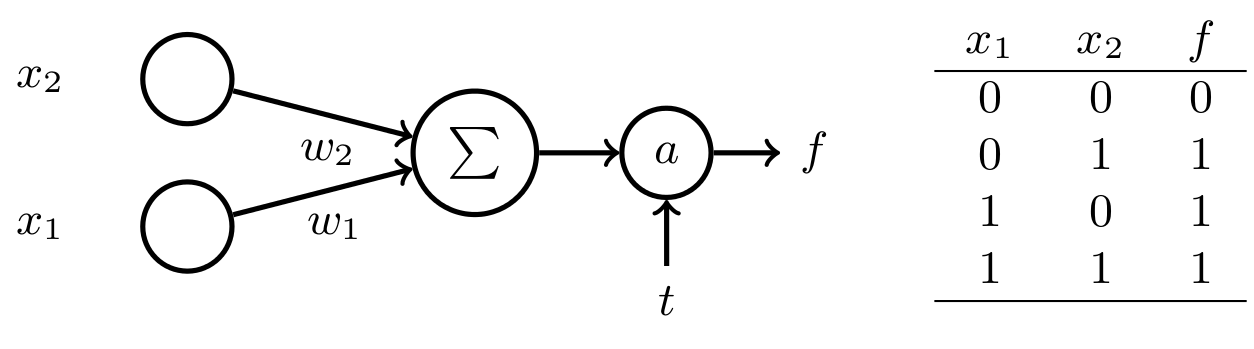}
\caption{MCP y tabla de verdad (OR).}
\label{fig:mcp-or}
\end{figure}

Para mostrar el funcionamiento básico de una neurona artificial (MCP) en la \ref{fig:mcp-or} se muestra el MCP para una compuerta lógica OR con dos entradas \cite{Aleksander:1992:book}.

  Sabemos que $f=1$, si $\sum_{i=1}^{n}x_i w_i > t$, así que si sustituimos los valores de $x_i$ de la tabla de verdad (\ref{fig:mcp-or}) obtenemos las siguientes desigualdades: \\
 \begin{tabular}{ll}
   $0(w_1)+0(w_2)<t$ & $0 < t$\\
   $0(w_1)+1(w_2)>t$ & $w_2>t$\\
   $1(w_1)+0(w_2)>t$ & $w_1>t$\\
   $1(w_1)+1(w_2)>t$ & $w_1+w_2>t$\\
 \end{tabular} \\
    
Se recomienda asignar valores a los pesos $w_1, w_2$ en el rango $[-1,+1]$. Posibles valores que cumplen con las desigualdades son $t = 0.5$; $w_1=0.7$; $w_2=0.7$. Es decir, para el problema de la compuerta lógica OR de dos entradas $x_1$ y $x_2$ (\ref{fig:mcp-or}), la neurona que simula el comportamiento de $f$ está dado por los pesos $w_1$ y $w_2$ y el umbral $t$.
 
 \subsubsection{Interpretación gráfica: clasificador lineal}

Si reemplazamos los valores $t = 0.5$, $w_1=0.7$ y $w_2=0.7$ (OR) en el MCP $\sum_{i=1}^{n}x_i w_i > t$, obtenemos: \\

\noindent$0.7x_1+0.7x_2 = 0.5$ \\
$x_2 = (0.5 - 0.7x_1)/0.7$\\
$x_2 = 0.5/0.7 - x_1$ \\

\begin{figure}[!ht]
\includegraphics[width=2.5in]{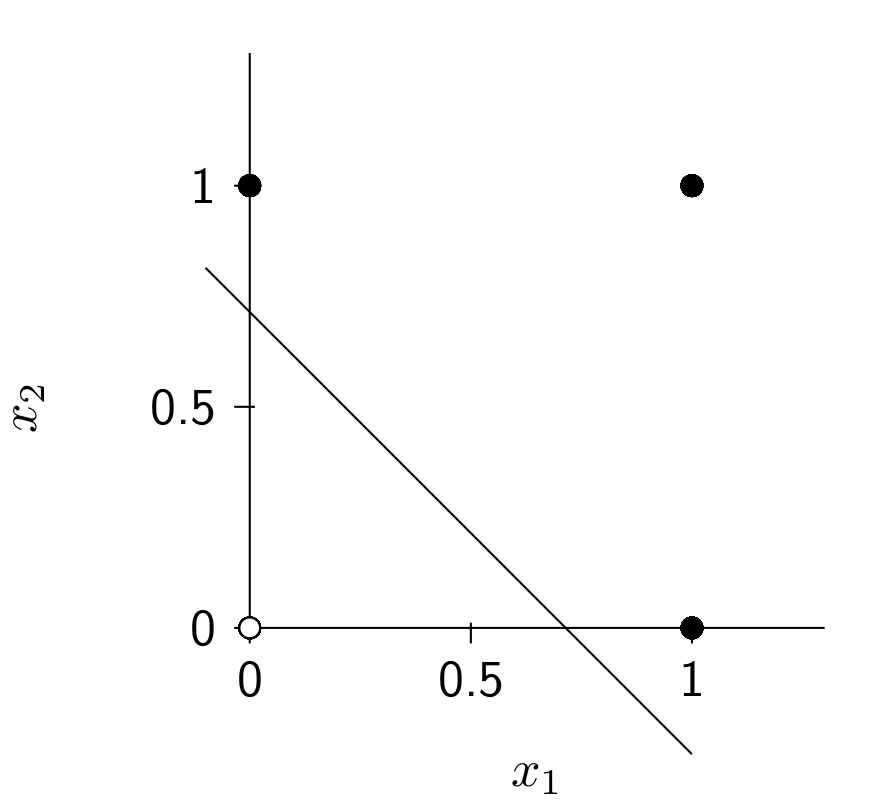}
\caption{MCP: Clasificador lineal.}
\label{fig:graph-mcp-or}
\end{figure}

Esto representa la ecuación de una recta. Si $x_1=0$, obtenemos el punto de corte de la recta en $x_2=0.71$. Ver recta de clasificación en la \ref{fig:graph-mcp-or}.

En la \ref{fig:code-mcp-or} se muestra el código de un script en MATLAB\textsuperscript{\textregistered{}}/Octave para graficar la recta de la \ref{fig:graph-mcp-or}.

\begin{figure}[!ht]
 \begin{lstlisting}
% MATLAB/Octave script
% MCP OR
close all;
% Tabla de verdad de compuerta OR 
x = [0 0 1 1;
     0 1 0 1];    % Entradas, datos de entrenamiento
f = [0 1 1 1];    % Salida deseada (target)
tam = size(f,2);  % # de muestras
figure;
hold on;
for i=1:tam,
     if(f(i) == 1),
  	plot(x(1,i),x(2,i),'k*');
      else
 	plot(x(1,i),x(2,i),'ko');
      end  
end
x1=-0.1:1.1,
x2=0.5/0.7 - x1; % ec. para obtener la recta 			
plot(x1,x2,'b');
xlabel('x_1');
ylabel('x_2');
title('MCP OR')
grid on;
 \end{lstlisting}
 \caption{MATLAB\textsuperscript{\textregistered{}}/Octave script: Ejemplo MCP, compuerta lógica OR de dos entradas.}
 \label{fig:code-mcp-or}
\end{figure}

\subsection{Ejemplo: 3 entradas}
\begin{figure}
\begin{tabular}{@{}m{3.5in}@{} @{}m{1.5in}@{}} 
\includegraphics[width=3in]{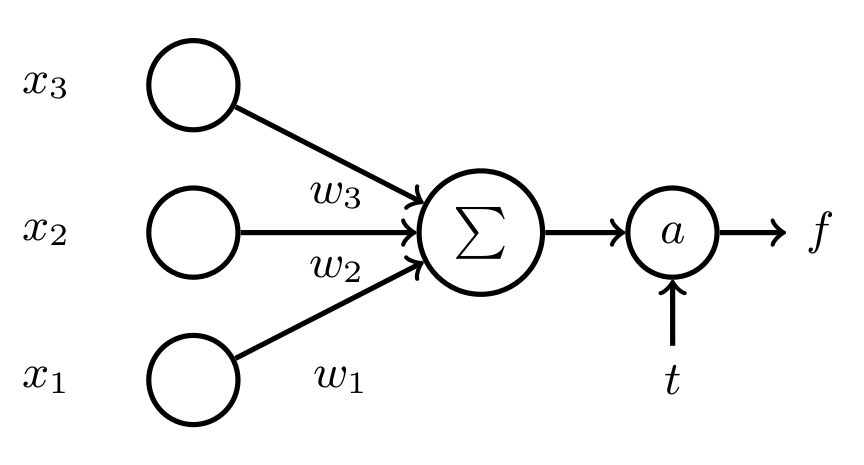} &
\begin{tabular}{cccc}
$x_1$ & $x_2$ & $x_3$ & $f$ \\ \hline
0 & 0 & 0 & 1 \\
0 & 0 & 1 & 1 \\
0 & 1 & 0 & 0 \\
0 & 1 & 1 & 0 \\
1 & 0 & 0 & 0 \\
1 & 0 & 1 & 1 \\
1 & 1 & 0 & 0 \\
1 & 1 & 1 & 0 \\ \hline
\end{tabular}
\end{tabular}
\caption{Ejemplo de tres entradas}
\label{fig:3-entradas}
\end{figure}

En la \ref{fig:3-entradas} se muestra un ejemplo de una neurona (MCP) con tres entradas.

Las desigualdades de acuerdo al MCP son: \\
 \begin{tabular}{ll}
   $0(w_1)+0(w_2)+0(w_3)>t$ & $0 > t$\\
   $0(w_1)+0(w_2)+1(w_3)>t$ & $w_3>t$\\
   $0(w_1)+1(w_2)+0(w_3)<t$ & $w_2<t$\\
   $0(w_1)+1(w_2)+1(w_3)<t$ & $w_2+w_3<t$\\
   $1(w_1)+0(w_2)+0(w_3)<t$ & $w_1<t$\\
   $1(w_1)+0(w_2)+1(w_3)>t$ & $w_1+w_3>t$\\
   $1(w_1)+1(w_2)+0(w_3)<t$ & $w_1+w_2<t$\\
   $1(w_1)+1(w_2)+1(w_3)<t$ & $w_1+w_2+w_3<t$\\
 \end{tabular} \\

\begin{figure}[!ht]
\includegraphics[width=4.5in]{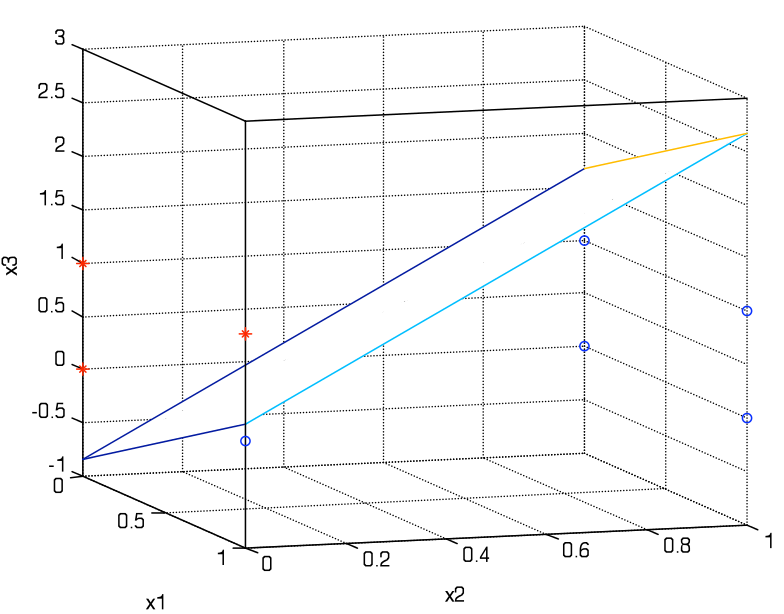} 
\caption{Hiperplano clasificador de un MCP de tres entradas, ver \ref{fig:3-entradas}.}
\label{fig:3-entradas-hiperplano}
\end{figure}

Un conjunto posible de valores para los pesos es \{$w_1=-0.6; w_2=-1.5; w_3=0.6; t=-0.5$\}. El hiperplano para el MCP de tres entradas está dado por $x_1w_1+x_2w_2+x_3w_3 = t$, ver la \ref{fig:3-entradas-hiperplano}.

El código para visualizar el hiperplano de la \ref{fig:3-entradas} se muestra en la \ref{fig:code-mcp-3-entradas}.

\begin{figure}[!ht]
 \begin{lstlisting}
% MATLAB/Octave script
clear;
hold on;
view(72,10);
[x,y]=meshgrid(0:1,0:1);
T=-0.5;w3=0.6;w2=-1.5;w1=-0.6;
z=(T-w1.*x-w2.*y)/w3;
mesh(x,y,z);
plot3(0,0,0,'r*');
plot3(0,0,1,'r*');
plot3(0,1,0,'bo');
plot3(0,1,1,'bo');
plot3(1,0,0,'bo');
plot3(1,0,1,'r*');
plot3(1,1,0,'bo');
plot3(1,1,1,'bo');
xlabel('x1');
ylabel('x2');
zlabel('x3');
plot3([1 1],[0 0],[-1 3],'k-');
plot3([0 1],[0 0],[3 3],'k-');
plot3([1 1],[0 1],[3 3],'k-');
grid on;
 \end{lstlisting}
 \caption{MATLAB\textsuperscript{\textregistered{}}/Octave script: MCP ejemplo de tres entradas.}
 \label{fig:code-mcp-3-entradas}
\end{figure}

\section{Aprendizaje: Regla Delta}

En la sección anterior se describió el MCP. Los pesos de una neurona se encuentran a través de las desigualdades. Es un proceso manual que si extendemos los ejemplos a $4,5,\dots,n$ entradas, el número de desigualdades crece exponencialmente.

Por lo anterior no resulta práctico el MCP sin un algoritmo que permita encontrar los pesos de manera automática.

En 1962, Bernard Widrow y Ted Hoff propusieron la regla delta, un algoritmo iterativo (ver Algoritmo \ref{alg}) para obtener los pesos de una neurona iterativamente (automáticamente), también conocida como regla de aprendizaje \cite{Widrow-Hof:1962:RD,Aleksander:1992:book}.

\begin{algorithm}[H]
\caption{Regla Delta}
\label{alg}
Selecciona una entrada de la tabla de verdad\;
Si se detecta un error, estimar que tan lejos está el MCP de la salida deseada \;
Ajustar los pesos `activos' (i.e. cuando una entrada $x_i=1$) para remover la porción $d=(E+e)/2$ del error\;
Ir al paso 1, hasta que ninguna columna genere un error\;
\end{algorithm}

El factor de corrección $d$ se define por el error $E$ y una constante de aprendizaje $e$ (similar al valor de temperatura en recocido simulado -- simulated annealing) \cite{Rich:1994:book}.

\subsection{Ejemplo: neurona de 2 entradas}
\begin{figure}[!ht]
\includegraphics[width=4in]{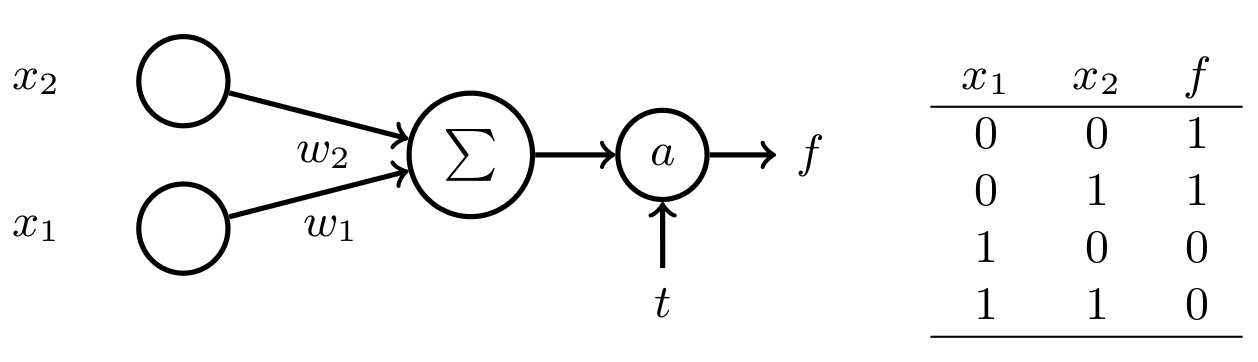}
\caption{Ejemplo regla delta de 2 entradas para la Regla Delta.}
\label{fig:regla-delta-2-entradas}
\end{figure}

En la \ref{fig:regla-delta-2-entradas} se muestra un ejemplo de neurona (MCP) de dos entradas \cite{Aleksander:1992:book}, con este ejemplo se ejecutará el Algoritmo \ref{alg} (Regla Delta).

Recordar que el MCP se define como $\sum_{i=1}^{n}x_i w_i > t$, cuando una neurona se activa $f=1$. Los pesos iniciales para el ejemplo de la regla delta son:  $w_1=0.2; w_2=-0.5; t=0.1$. Los valores iniciales se definen de manera aleatoria (random), preferentemente en un rango de $[-1,+1]$. El valor de la constante de aprendizaje se establece a $e=0.1$.

\begin{table}
\caption{Ejecución del Algoritmo \ref{alg}: Regla Delta}
\label{tbl:regla-delta}
\begin{tabular}{clll|l|l|l|}
      Iteración & Columna & Error & Corrección & Nuevo & Nuevo & Nuevo \\
       & $x_1, x_2$ & $E$ & $d=(E+e)/2$ & $w_1$ & $w_2$ & $t$ \\ \hline
      1 & 0,0 & 0.1 & 0.1 & -- & -- & $0$ \\
      2 & 0,1 & 0.5 & 0.3 & -- & $-0.2$ & $-0.3$ \\
      3 & 1,1 & 0.3 & 0.2 & $0$ & $-0.4$ & $-0.1$ \\
      4 & 1,0 & 0.1 & 0.1 & $-0.1$ & -- & $0$ \\
      5 & 0,1 & 0.4 & 0.25 & -- & $-0.15$ & $-0.25$ \\
      6 & 1,0 & 0.15 & 0.125 & $-0.225$ & -- & $-0.125$ \\
      7 & 0,1 & 0.025 & 0.0625 & -- & $-0.0875$ & $-0.1875$ \\
       \hline
   \end{tabular}  
\end{table}

Cada iteración en la \ref{tbl:regla-delta} corresponde a la ejecución de los tres pasos del Algoritmo \ref{alg}. 

\begin{itemize}
\item Iteración 1
\begin{enumerate}
\item Se selecciona la entrada $0,0$ de la tabla de verdad en la \ref{fig:regla-delta-2-entradas}.
\item Los valores de $x_1, x_2$ se sustituyen en el MCP, $0(w_1)+0(w_2)>t$, porque $f=1$. Con los valores iniciales $w_1=0.2, w_2=-0.5, t=0.1$, la desigualdad resultante es $0 > 0.1$. Por lo tanto el error es $E=0.1$, porque no se satisface la desigualdad.
\item Los pesos `activos', son solo $t$, dado que $x_1=0$ y $x_2=0$. Se aplica el factor de corrección $d=(E+e)/2=(0.1+0.1)/2=0.1$ al valor de $t$. El razonamiento de la corrección es el siguiente: $\mathbf{+}>\mathbf{-}$ y $\mathbf{-}<\mathbf{+}$, es decir a los pesos se les suma o se les resta la corrección dependiendo del tipo de desigualdad ($>$ o $<$). En este caso, se le resta $d$ a $t$, por lo tanto el nuevo valor de $t=t-d=0.1-0.1=0$ (ver \ref{tbl:regla-delta}).
\end{enumerate}
\item Iteración 2
\begin{enumerate}
\item Se selecciona la entrada $0,1$ de la tabla de verdad en la \ref{fig:regla-delta-2-entradas}.
\item Los valores de $x_1, x_2$ se sustituyen en el MCP, $0(w_1)+1(w_2)>t$, porque $f=1$. Con los valores actuales $w_1=0.2, w_2=-0.5, t=0$, la desigualdad resultante es $-0.5 > 0$. Por lo tanto el error es $E=0.5$, porque no se satisface la desigualdad.
\item Los pesos `activos', son $w_2$ y $t$, dado que $x_1=0$ y $x_2=1$. Se aplica el factor de corrección $d=(E+e)/2=(0.5+0.1)/2=0.3$ a $w_2$ y $t$. En este caso, se le suma $d$ a $w_2$ y se le resta $d$ a $t$ (ver Iteración 1), por lo tanto los nuevos valores son $w_2=w_2-d=-0.5+0.3=-0.2$ y  $t=t-d=0-0.3=-0.3$ (ver \ref{tbl:regla-delta}).
\end{enumerate}
\item Iteración 3
\begin{enumerate}
\item Se selecciona la entrada $1,1$ de la tabla de verdad en la \ref{fig:regla-delta-2-entradas}.
\item Los valores de $x_1, x_2$ se sustituyen en el MCP, $1(w_1)+1(w_2)<t$, porque $f=0$. Con los valores actuales $w_1=0.2, w_2=-0.2, t=-0.3$, la desigualdad resultante es $0.2+(-0.2) < -0.3$, es decir $0 < -0.3$. Por lo tanto el error es $E=0.3$, porque no se satisface la desigualdad.
\item Los pesos `activos', son $w_1$, $w_2$ y $t$, dado que $x_1=1$ y $x_2=1$. Se aplica el factor de corrección $d=(E+e)/2=(0.3+0.1)/2=0.2$ a $w_1$, $w_2$ y $t$. En este caso, se le resta $d$ a $w_1$, $w_2$ y se le suma $d$ a $t$ (ver Iteración 1), por lo tanto los nuevos valores son $w_1=w_1-d=0.2-0.2=0$, $w_2=w_2-d=-0.2-0.2=-0.4$ y $t=t+d=-0.3+0.2=-0.1$ (ver \ref{tbl:regla-delta}).
\end{enumerate}
\item $\dots$ $\dots$
\item Iteración 7
\begin{itemize}
\item La \ref{tbl:regla-delta} termina en la Iteración 7, porque los pesos obtenidos al finalizar la iteración ya no generan error. Las desigualdades para la tabla de verdad en la \ref{fig:regla-delta-2-entradas} son:\\
 \begin{tabular}{lll}
   $0(w_1)+0(w_2)>t$ & $0 > t$ & $0 > -0.1875$\\
   $0(w_1)+1(w_2)>t$ & $w_2>t$ & $-0.0875 > -0.1875$\\
   $1(w_1)+0(w_2)<t$ & $w_1<t$ & $-0.225 < -0.1875$\\
   $1(w_1)+1(w_2)<t$ & $w_1+w_2<t$ & $(-0.0875-0.225)<-0.1875$\\
 \end{tabular} \\
 Todas las desigualdades se cumplen, es decir $E=0$, se cumple el paso 4 del Algoritmo \ref{alg} y termina.
\end{itemize}
\end{itemize}

\section{Problemas no-lineales}
Una neurona (MCP) funciona para la solución de problemas lineales, tal como se ilustra gráficamente en \ref{fig:graph-mcp-or} y \ref{fig:3-entradas}. Lo mismo para la Rela Delta del Algoritmo \ref{alg}, funciona para problemas lineales. 

El problema del XOR es un problema no-lineal que una neurona (MCP) no puede resolver. 

\subsection{Ejemplo: XOR de 2 entradas}
\begin{figure}[!ht]
\includegraphics[width=4in]{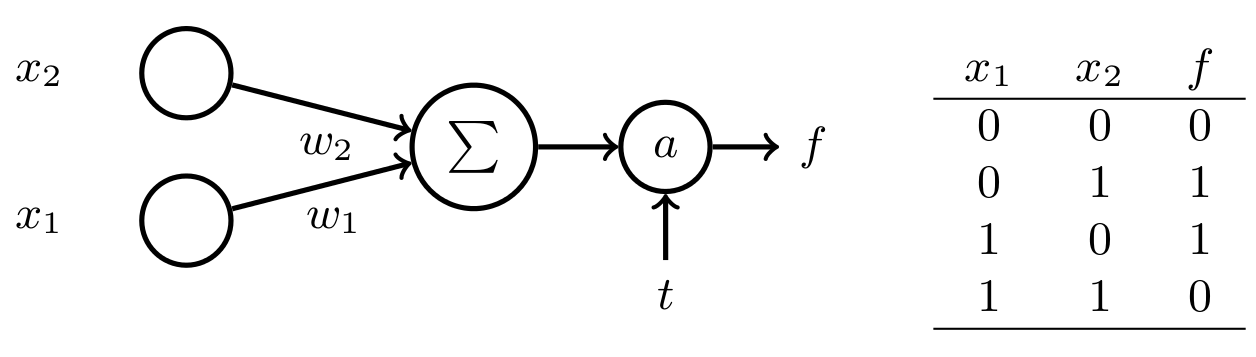}
\caption{Neurona de dos entradas y tabla de verdad (XOR).}
\label{fig:xor-truth-table}
\end{figure}

\begin{figure}[!ht]
\includegraphics[width=2.5in]{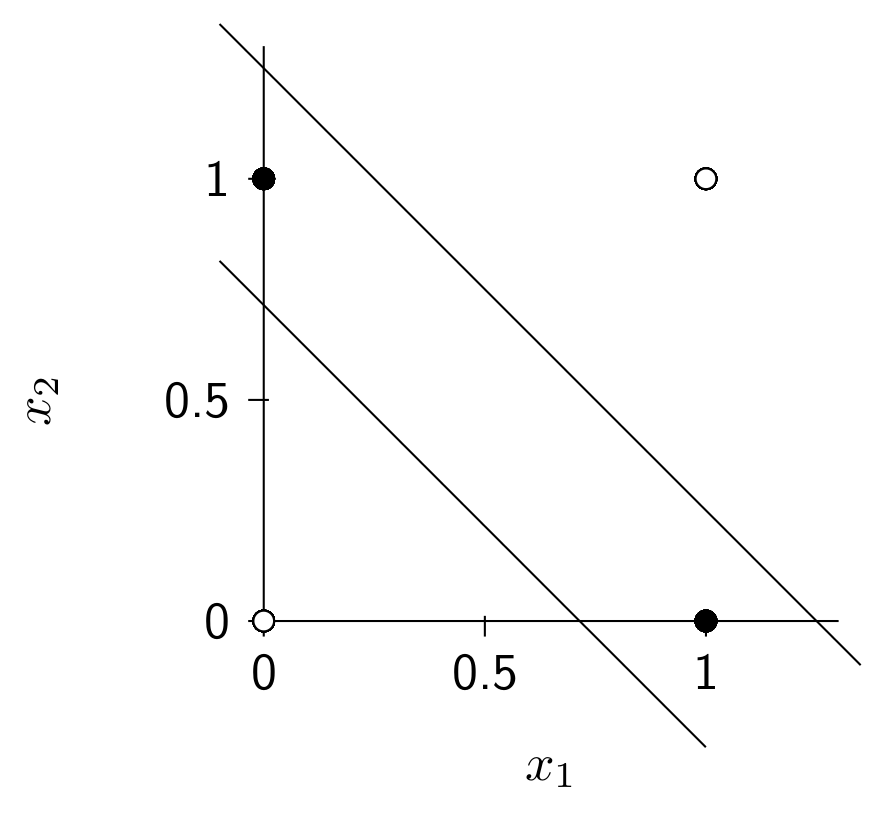}
\caption{Compuerta XOR de dos entradas.}
\label{fig:xor-graph}
\end{figure}

En la \ref{fig:xor-truth-table} se muestra la tabla de verdad de la compuerta XOR para dos entradas, y en la \ref{fig:xor-graph} se muestra gráficamente el problema XOR de dos entradas, claramente se ve que un clasificador lineal no resuelve el problema. Se requiere más de un clasificador lineal.

El MCP para las tabla de verdad de la \ref{fig:xor-truth-table} genera las siguientes desigualdades: \\
 \begin{tabular}{ll}
   $0(w_1)+0(w_2)<t$ & $0 < t$\\
   $0(w_1)+1(w_2)>t$ & $w_2>t$\\
   $1(w_1)+0(w_2)>t$ & $w_1>t$\\
   $1(w_1)+1(w_2)<t$ & $w_1+w_2<t$\\
 \end{tabular} \\

Si $t$ es positivo, por ejemplo $t=1$, entonces $w_1=2$ y $w_2=2$. Se cumplen las tres primeras desigualdades, $0<1$, $2 > 1$ y $2 >1$; pero no se cumple $4<1$, porque es una contradicción que individualmente $w_1>t$ y $w_2>t$ y después juntas sean menor a $t$. Esta es una forma de demostrar matemáticamente a través del MCP y las desigualdades cuando un problema es no-lineal. Gráficamente se puede desmostrar a través de la \ref{fig:xor-graph}. Sin embargo, cuando tenemos $4,5,\dots,n$ variables es imposible gráficarlo, el único camino es através de las desigualdades.

\subsection{Ejemplo: 3 entradas}
\begin{figure}
\begin{tabular}{@{}m{3.5in}@{} @{}m{1.5in}@{}} 
\includegraphics[width=3in]{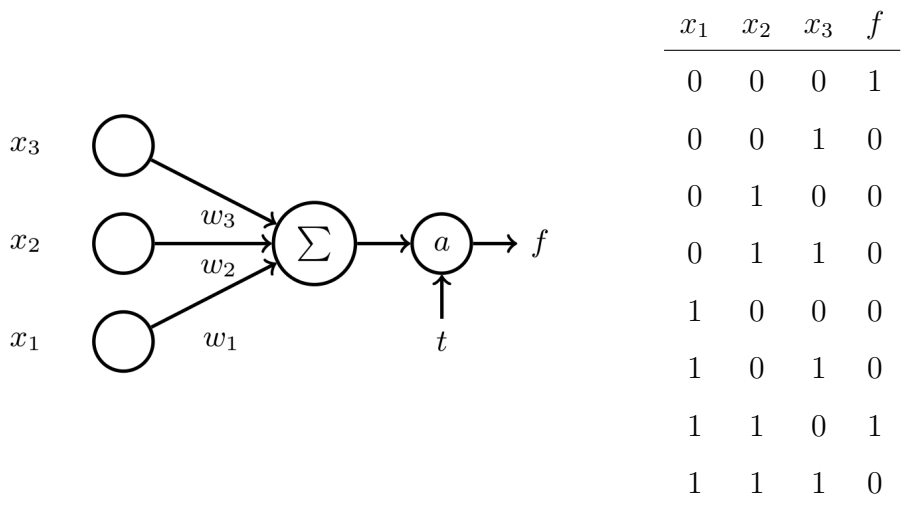} &
\begin{tabular}{cccc}
$x_1$ & $x_2$ & $x_3$ & $f$ \\ \hline
0 & 0 & 0 & 1 \\
0 & 0 & 1 & 0 \\
0 & 1 & 0 & 0 \\
0 & 1 & 1 & 0 \\
1 & 0 & 0 & 0 \\
1 & 0 & 1 & 0 \\
1 & 1 & 0 & 1 \\
1 & 1 & 1 & 0 \\ \hline
\end{tabular}
\end{tabular}
\caption{Ejemplo de tres entradas: no-lineal}
\label{fig:3-entradas-xor}
\end{figure}

En la \ref{fig:3-entradas-xor} se muestra un ejemplo de una neurona (MCP) con tres entradas así como la tabla de verdad para un problema no-lineal.

Las desigualdades de acuerdo al MCP son: \\
 \begin{tabular}{ll}
   $0(w_1)+0(w_2)+0(w_3)>t$ & $0 > t$\\
   $0(w_1)+0(w_2)+1(w_3)<t$ & $w_3<t$\\
   $0(w_1)+1(w_2)+0(w_3)<t$ & $w_2<t$\\
   $0(w_1)+1(w_2)+1(w_3)<t$ & $w_2+w_3<t$\\
   $1(w_1)+0(w_2)+0(w_3)<t$ & $w_1<t$\\
   $1(w_1)+0(w_2)+1(w_3)<t$ & $w_1+w_3<t$\\
   $1(w_1)+1(w_2)+0(w_3)>t$ & $w_1+w_2>t$\\
   $1(w_1)+1(w_2)+1(w_3)<t$ & $w_1+w_2+w_3<t$\\
 \end{tabular} \\

Como $t$ tiene que ser menor a cero, además $w_1 < t$ y $w_2<t$.
Por lo tanto la desigualdad $w_1+w_2>t$ nunca será satisfecha, es decir es un problema linealmente no separable, no-lineal.

\begin{figure}[!ht]
\includegraphics[width=4.5in]{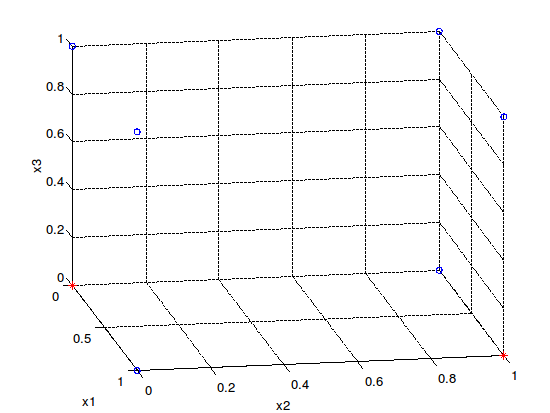} 
\caption{Problema no-lineal de tres entradas, ver tabla de verdad en la \ref{fig:3-entradas-xor}.}
\label{fig:3-entradas-no-lineal}
\end{figure}

El problema no-lineal de tres entradas de la \ref{fig:3-entradas-xor} se puede ver gráficamente en la \ref{fig:3-entradas-no-lineal}. Para lograr la separación se requiere un plano curvo (no-lineal) que cruce dos veces el plano $x_1,x_2$, de tal forma que los puntos ($x_1=1,x_2=0$) y ($x_1=1,x_2=1$) (con $x_3=0$) queden por debajo del plano curvo y los demás puntos por arriba.

El código para visualizar  la \ref{fig:3-entradas-no-lineal} se muestra en la \ref{fig:code-mcp-3-entradas-no-lineal}.

\begin{figure}[!ht]
 \begin{lstlisting}
% MATLAB/Octave script
% Problema1 capitulo II, F4
hold on;
view(80,20);
plot3(0,0,0,'r*');
plot3(0,0,1,'bo');
plot3(0,1,0,'bo');
plot3(0,1,1,'bo');
plot3(1,0,0,'bo');
plot3(1,0,1,'bo');
plot3(1,1,0,'r*');
plot3(1,1,1,'bo');
xlabel('x1');
ylabel('x2');
zlabel('x3');
grid on;
 \end{lstlisting}
 \caption{MATLAB\textsuperscript{\textregistered{}}/Octave script: ejemplo no-lineal de tres entradas.}
 \label{fig:code-mcp-3-entradas-no-lineal}
\end{figure}

\subsection{Perceptron}

El problema del XOR es un problema no-lineal que un clasificador simple no puede resolver. En la literatura de redes neuronales artificiales, una neurona o \emph{perceptron}, es el modelo matemático de una neurona más estudiado. El \emph{perceptron} fue introducido por  Frank Rosenblatt en 1957~\cite{Rosenblatt:1957:perceptron}. Una década después, Marvin Minsky y Seymour Paper escribieron el famoso libro: \emph{Perceptrons}. Ellos desmostraron que el perceptron no puede resolver el problema XOR \cite{Minsky:1969:book}.

Quizás esa publicación hizo que la investigación en el área de redes neuronales artificiales se detuviera hasta que apareció el algoritmo backpropagation~\cite{Rumelhart:1986:Backpropagation,Bishop:1995:book,Rojas:1996:book,Haykin:1999:book}, casi 20 años después de la publicación de Minsky y Paper \cite{Minsky:1969:book}.

\section{Redes Neuronales: Backpropagation}
\begin{figure}
\includegraphics[width=3.5in]{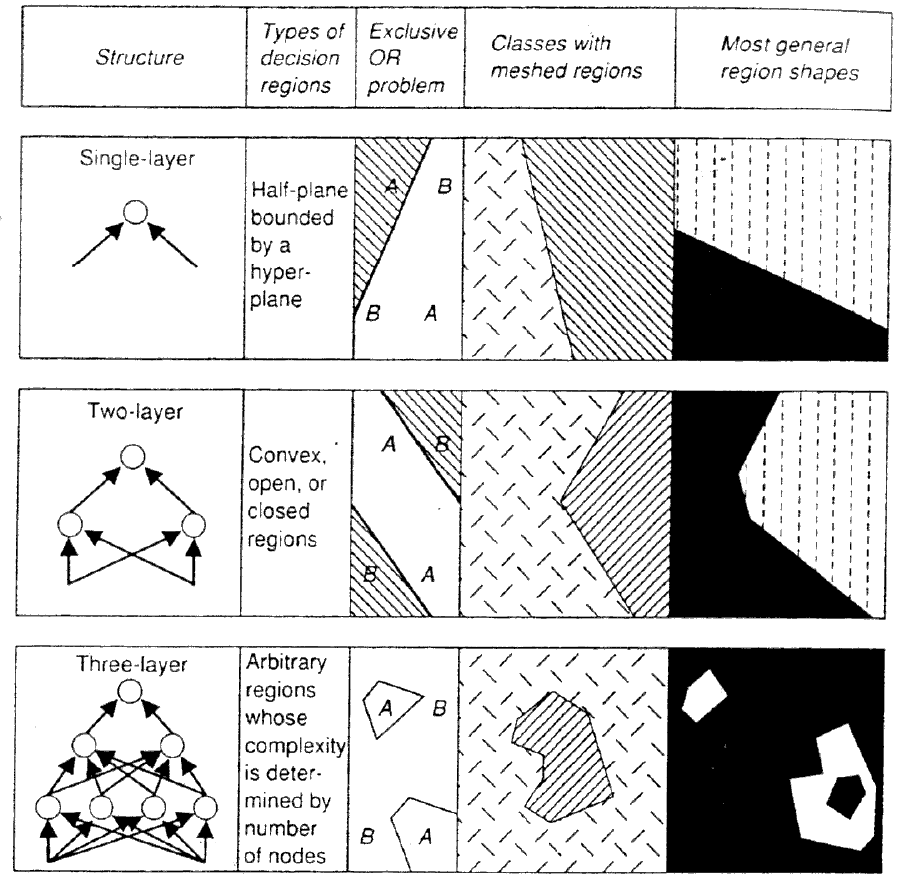}
\caption{Capacidad de clasificación de las redes neuronales artificiales según su dimensión \cite{Lippmann:1987:BPN}.}
\label{fig:decision-regions}
\end{figure}

La complejidad o capacidad de clasificación de las redes neuronales artificiales depende de las cantidad de neuronas, ver la \ref{fig:decision-regions}. Una neurona (una capa -- one layer) es un clasificador lineal, una red con estructura de dos o más capas (layers) permite resolver problemas no-lineales.

El algoritmo de backpropagation es la generalización de la Regla Delta \cite{Rojas:1996:book}, descrita anteriormente. También se conoce como perceptron multicapa (Multilayer Perceptrons -- MLP), red neuronal de retropropagación (Feed Forward Neural Networks -- FFNN) \cite{Rumelhart:1986:Backpropagation,Bishop:1995:book,Haykin:1999:book}.

\subsection{Ejemplo XOR de dos entradas} 
\begin{figure}
\includegraphics[width=4.5in]{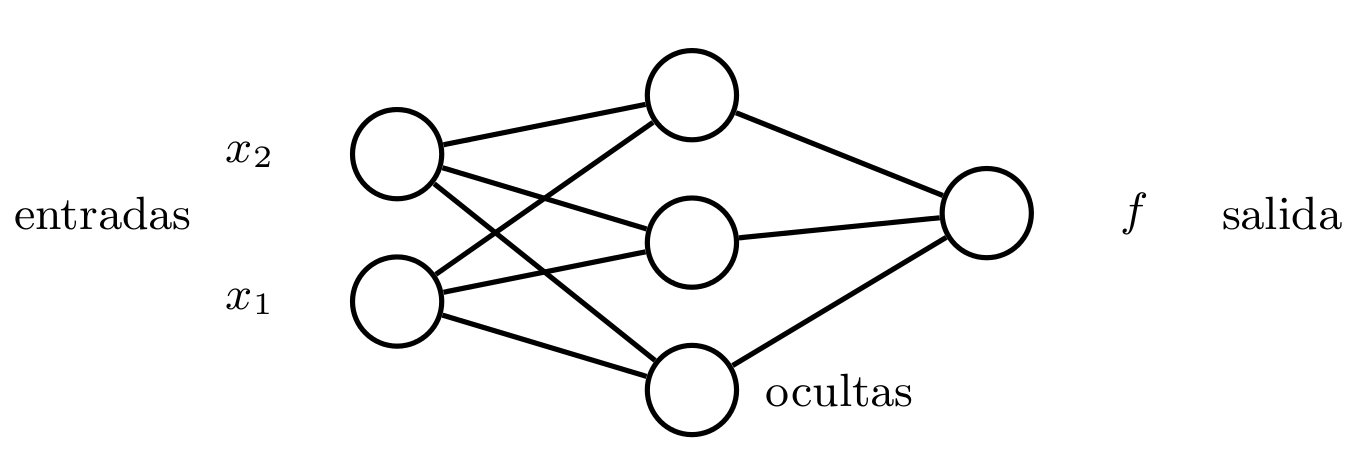}
\caption{Arquitectura de red neuronal artificial, tipo backpropagation. Tiene dos entradas, una capa oculta y una salida.}
\label{fig:xor-bpn}
\end{figure}

En la \ref{fig:xor-bpn} se muestra una arquitectura de red neuronal que resuleve el problema de XOR de dos entradas.

En la \ref{fig:code-xor-bpn} se muestra el código en MATLAB\textsuperscript{\textregistered{}} para resolver el problema de XOR de dos entradas.

\begin{figure}
 \begin{lstlisting}
 % MATLAB script
 p = [0 0 1 1;0 1 0 1];  % Inputs
 t = [0 1 1 0];          % Output (target)
 net=newff(minmax(p),[3,1],{'tansig','purelin'},'traingd'); % Create FFNN
 [net,tr]=train(net,p,t); % Training
 a = sim(net,p)           % Testing
 % a = 0.0034    0.9962    0.9942    0.0028
 \end{lstlisting}
\caption{Código MATLAB para resolver el problema de XOR de dos entradas con backpropagation.}
\label{fig:code-xor-bpn}
\end{figure}

\subsection{Ejemplo: clasificador de figuras}

\begin{figure}
\includegraphics[width=2.5in]{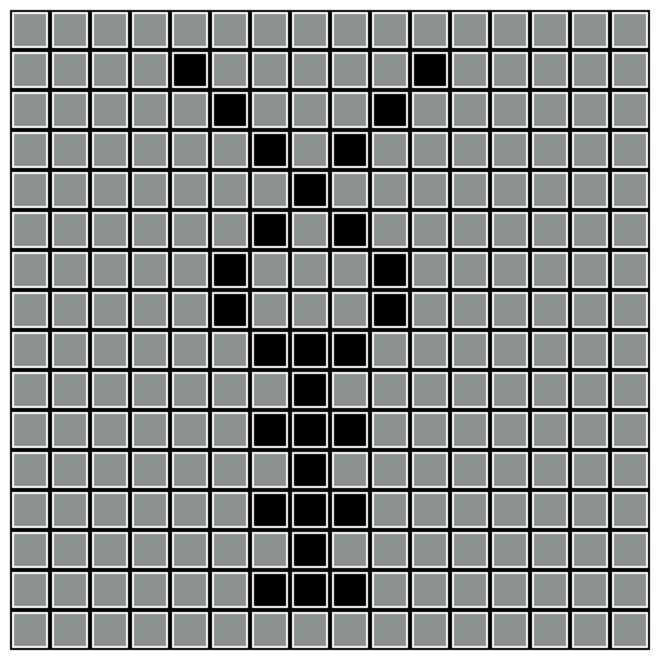}
\caption{Ejemplo de imagen de $16 \times 16$ pixeles.}
\label{fig:imagen-16-x-16}
\end{figure}

En la \ref{fig:imagen-16-x-16} se muestra una imagen (figura) de $[16 \times 16]$. El objetivo es crear un clasificador de imagenes a través de una red neuronal artificial, backpropagation.

\begin{figure}
\includegraphics[width=4.5in]{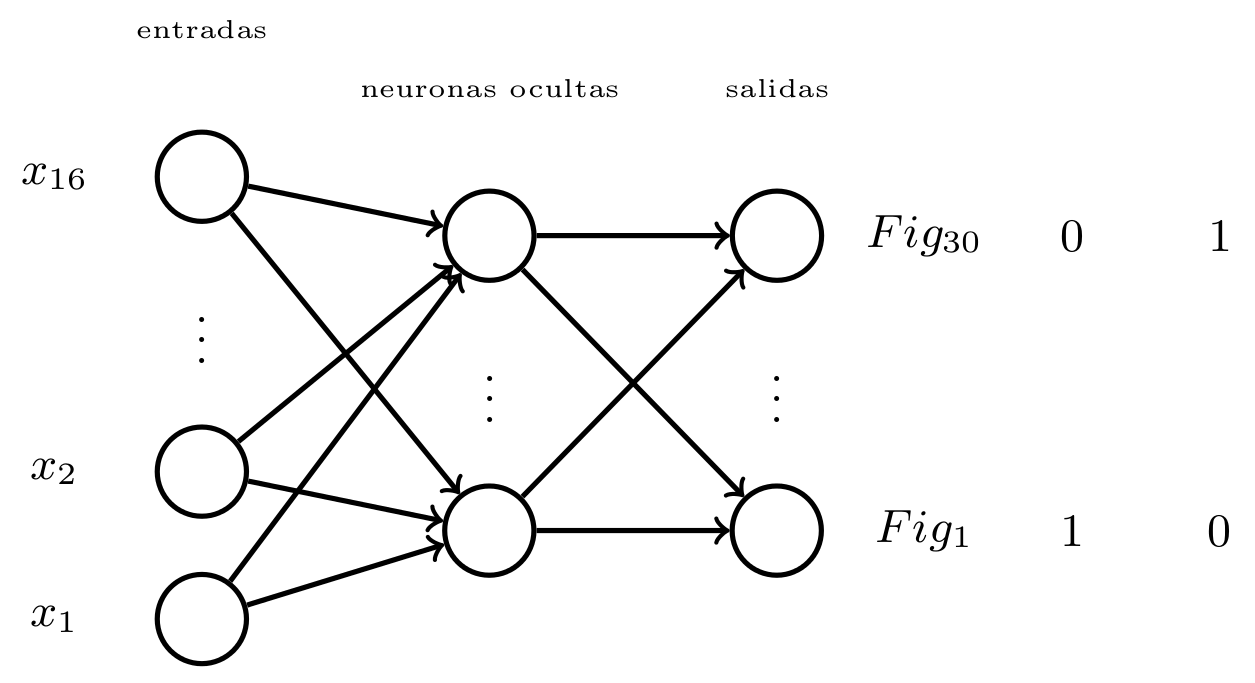}
\caption{Red neuronal artificial para clasificar imagenes de $16 \times 16$ pixeles.}
\label{fig:BPN-imagenes-16-x-16}
\end{figure}

La arquitectura de red neuronal artificial en la \ref{fig:BPN-imagenes-16-x-16} contiene: 16 entradas,
32 neuronas en la capa oculta y 30 salidas. El número de salidas determina la cantidad de imagenes que puede clasificar esta red. En cada salida se tiene un valor entre [0,1] el valor de umbral para decidir si es una imagen/figura válida es de 0.8 (threshold).

\section{Comentarios finales}
En estos apuntes se enfocan únicamente al funcionamiento de una neurona artificial, el modelo de McCulloch \& Pitts (MCP), la regla delta y backpropagation.

Sin embargo, es importante resaltar que existen diversas tecnologías de redes neuornales artificiales y algoritmos de clasificación:
\begin{itemize}
 \item Deep Neural Networks (DNN) - iterativo
 \item Covolutional Neural Networks (CNN) - iterativo
 \item Probabilistic Neural Networks (PNN) - determinístico
 \item Support Vector Machines (SVM) - determinístico
 \item Métodos Bayesianos
 \item Entre otros
\end{itemize}

\section*{Agradecimientos}
Se agradece a Rogelio Fernando Cabañes, estudiante de movilidad de la Universidad Autónoma del Estado de México, por la revisión de los apuntes y sus observaciones. También se agradece al programa ``Estímulos al Desempeño del Personal Docente'' de la UASLP, que gracias al programa fue posible la elaboración de estos apuntes.

\bibliographystyle{plain}
\bibliography{references.bib}

\begin{thebibliography}{10}

\bibitem{Aleksander:1992:book}
I.~Aleksander and E.~Horton.
\newblock {\em An Introduction to Neural Computing}.
\newblock Chapmand \& Hal, 1992.

\bibitem{Bishop:1995:book}
C.~M. Bishop and G.~E. Hinton.
\newblock {\em Neural Networks for Pattern Recognition}.
\newblock Clarendon Press, 1995.

\bibitem{Haykin:1999:book}
S.~Haykin.
\newblock {\em Neural Networks: {A} Comprehensive Foundation}.
\newblock Prentice Hall, 1999.

\bibitem{Lippmann:1987:BPN}
R.~Lippmann.
\newblock An introduction to computing with neural nets.
\newblock {\em IEEE Magazine}, 4(2):4--22, 1987.

\bibitem{Minsky:1969:book}
M.~L. Minsky and S.~A. Papert.
\newblock {\em Perceptrons}.
\newblock Cambridge, MA: MIT Press, 1969.

\bibitem{Rich:1994:book}
E.~Rich and K.~Knight.
\newblock {\em Inteligencia {A}rtificial}.
\newblock Mc Graw Hill, 1994.

\bibitem{Rojas:1996:book}
R.~Rojas.
\newblock {\em Neural Networks: {A} Systematic Introduction}.
\newblock Springer-Verlag, 1996.

\bibitem{Rosenblatt:1957:perceptron}
F.~Rosenblatt.
\newblock The perceptron--{A} perceiving and recognizing automaton.
\newblock Technical Report 85-460-1, Cornell Aeronautical Laboratory, 1957.

\bibitem{Rumelhart:1986:Backpropagation}
D.~E. Rumelhart, G.~E. Hinton, and R.J. Williams.
\newblock Learning representations by back-propagating errors.
\newblock {\em Nature}, 323(1):533--536, 1986.

\bibitem{Widrow-Hof:1962:RD}
B.~Widrow and M.E. Hoff.
\newblock Associative {S}torage and {R}etrieval of {D}igital {I}nformation in
  {N}etworks of {A}daptive `neurons'.
\newblock {\em Biological Prototypes and Synthetic Systems}, 1:160, 1962.

\end{thebibliography}

\end{document}